\documentclass{article}
\pdfoutput=1
\PassOptionsToPackage{numbers, compress}{natbib}

\usepackage[final]{nips_2018}

\usepackage[utf8]{inputenc} 
\usepackage[T1]{fontenc}    
\usepackage{hyperref} 
\usepackage{url}            
\usepackage{booktabs}       
\usepackage{amsfonts}       
\usepackage{nicefrac}       
\usepackage{microtype}      
\usepackage{xcolor}
\usepackage{graphicx}
\usepackage{mathtools}
\usepackage{amsmath}
\usepackage{tikz}
\usepackage[symbol]{footmisc}
\usepackage{natbib}

\DeclareMathOperator*{\argmax}{argmax}
\DeclareMathOperator*{\argmin}{argmin}

\newcommand{\mypm}{\mathbin{\smash{%
\raisebox{0.35ex}{%
            $\underset{\raisebox{0.5ex}{$\smash -$}}{\smash+}$%
            }%
        }%
    }%
}

\title{Improving Skin Condition Classification with a Question Answering Model}

\author{
  Mohamed Akrout\thanks{Research conducted while working at \href{www.triage.com}{Triage}}\\
  University of Toronto\\
  \texttt{makrout@cs.toronto.edu} \\
  \And
  Amir-massoud Farahmand \\
  Vector Institute \\
  \texttt{farahmand@vectorinstitute.ai} \\
  \AND
  Tory Jarmain \thanks{1, Adelaide St. E., Suite 3001, Toronto, M5C1J4, Canada} \\
  Triage \\
  \texttt{tory@triage.com} \\
}

\begin{document}

\maketitle

\begin{abstract}
We present a skin condition classification methodology based on a sequential pipeline of a pre-trained Convolutional Neural Network (CNN) and a Question Answering (QA) model. This method enables us to not only increase the classification confidence and accuracy of the deployed CNN system, but also enables  the  emulation  of  the  conventional  approach  of  doctors  asking  the relevant  questions in refining the ultimate diagnosis and differential. By combining the CNN output in the form of classification probabilities as a prior to the QA model and the image textual description, we greedily ask the best symptom that maximizes the information gain over symptoms. We demonstrate that combining the QA model with the CNN increases the accuracy up to 10\% as compared to the CNN alone, and more than 30\% as compared to the QA model alone.

\end{abstract}
\section{Introduction}
Convolutional Neural Network (CNN) classification algorithms have produced promising results in classifying skin lesions images. The work of Kawahara et al. \cite{kawahara2016deep} explores the idea of using a pre-trained ConvNet as a feature extractor to distinguish among 10 classes of non-dermoscopic skin images. Liao et al. \cite{liao2016deep} describe an attempt to construct a universal skin condition classification system by applying transfer learning \cite{yosinski2014transferable} on a deep CNN and fine-tuning its weights. Codella et al. \cite{codella2017deep} report state-of-the-art performance results using ConvNets to extract image descriptors by using a pre-trained model from the Imagenet Large Scale Visual Recognition Challenge (ILSVRC) 2012 \cite{russakovsky2015imagenet}. The performance of CNN based diagnosis has been shown to achieve the same accuracy level as a panel of expert dermatologists for both benign and malignant skin lesions \cite{esteva2017dermatologist}.
The limitation of these computer vision approaches to skin lesion classification, however, is that they do not interact with the user in order to improve the initial accuracy of the diagnosis. By answering relevant questions, the model can further narrow down the initial diagnosis that was based only on the image descriptions.\\
Such a QA system can be implemented through a decision tree based model \cite{kotsiantis2007supervised, sun2013learning, azari2002web}. The tree can be used to map an input vector (i.e. a sequence of questions) to one of the leaf nodes, which corresponds to a region in the state space (i.e. all possible diagnosis). In this work, we use the decision tree approach to build a QA model that takes the CNN output softmax probabilities as a prior and outputs a sequence of questions that the user answers based on the symptoms in the image description.\\
To our knowledge, this is the first time that the probabilities of an image classifier is combined with a QA model formulated as an inference problem.

\section{Methods}
Our method combines a CNN classification model with a QA model that asks questions based on the classification probabilities of the CNN. We use the CNN conditions' output distribution as a prior in order to narrow down the relevant symptoms to be asked by the QA model.

\subsection{Classification of skin conditions}
\begin{figure}[h] 
\centering
\includegraphics[width=0.92\linewidth]{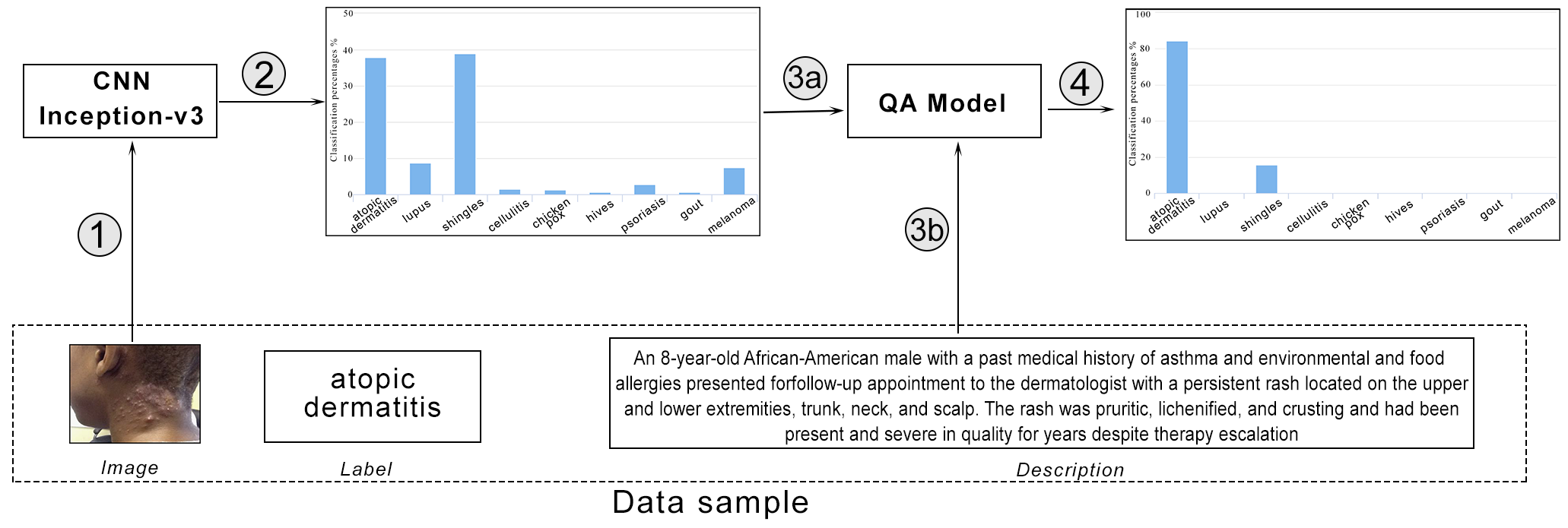} 
\caption{Classification pipeline during inference: \protect\tikz[inner sep=.25ex,baseline=-.75ex] \protect\node[circle,draw] {1}; and \protect\tikz[inner sep=.25ex,baseline=-.75ex] \protect\node[circle,draw] {2}; CNN Inference on the image, \protect\tikz[inner sep=.25ex,baseline=-.75ex] \protect\node[circle,draw] {3a}; Use the condition probabilities as a prior of the QA model, \protect\tikz[inner sep=.25ex,baseline=-.75ex] \protect\node[circle,draw] {3b}; Answer the QA questions based on the image description and \protect\tikz[inner sep=.25ex,baseline=-.75ex] \protect\node[circle,draw] {4}; compute the final condition probabilities } 
\label{fig:pipeline}
\end{figure}

We train an Inception-v3 network \cite{szegedy2016rethinking} on a dataset consisting of 5,841 images that are equally divided into 9 skin conditions: atopic dermatitis, lupus, shingles, cellulitis, chickenpox, hives, psoriasis, gout and melanoma.\\
As described in Figure~\ref{fig:pipeline}, we use annotated images (i.e. a labeled image with a paragraph describing the patient's symptoms), commonly called \textit{vignettes}, during inference in order to determine both the probability distribution over conditions of the CNN softmax output (step \tikz[inner sep=.25ex,baseline=-.75ex] \node[circle,draw] {2};) and the probability distribution over conditions updated by the QA model (step \tikz[inner sep=.25ex,baseline=-.75ex] \node[circle,draw] {4};). The labeled image is used by the CNN to classify the image (step \tikz[inner sep=.25ex,baseline=-.75ex] \node[circle,draw] {1};), and the symptoms mentioned in the image description are used to answer the QA system's questions (step \tikz[inner sep=.25ex,baseline=-.75ex] \node[circle,draw] {3b};). The image description does not include the true condition of the patient, but rather his symptoms.
\subsection{Inference combined with a QA model}
We formulate the QA sequential process as an inference problem where the condition distribution, initially provided by the CNN, is updated with each answered question. 
\subsubsection{Notation}
Let $\mathcal{S}$ and $\mathcal{C}$ be the symptom space and the condition space respectively where $|\mathcal{S}|=330$ and $|\mathcal{C}|=9$. We denote any symptom by $s_i$ and any condition $c_j$ where $1\leq j \leq 330$ and $1\leq i \leq 9$.

\subsubsection{The QA model}
To build the QA model, we assume that one chosen question to be asked corresponds to one symptom. We use the health knowledge matrix $\mathcal{M}$ \cite{rotmensch2017learning} of the 9 skin conditions learned from electronic medical records of 273,174 patients. The matrix $\mathcal{M}$ was evaluated and validated against both expert physician opinions and Google’s manually-constructed knowledge graph. It represents a matrix of condition-symptom relationships where each cell (i, j) represents the conditional probability $p(s_j | c_i)$ of a symptom $j$ given a condition $i$.
Using the matrix notation, we denote $\mathcal{M}$ by $p(S | C)$.\\\\
Using the CNN condition probabilities as the prior $p(C)$, we provide the QA model with the initial symptom $S_{init}$ of the patient from the image description. Thus, we can compute the updated condition probabilities $p(C|S)$ using the Bayes rule:
\begin{equation}
\label{eq:bayes-rule}
p(C|S) \propto P(S|C) \cdot P(C)
\end{equation}


Using the decision tree approach, we choose the best symptom to ask $s_j^*$ by maximizing the information gain (IG) over symptoms:
\begin{equation}
\begin{split}
s_j^* &= \argmax\limits_{s_j} \; IG(s_j,C)\\ &= \argmax\limits_{s_j} \big[H(C) - H(C|s_j)\big]\\ &=\argmin\limits_{s_j} H(C|s_j) \\ &=\argmax\limits_{s_j} \, \sum_{c} p(c|s_j)\; \log_2\;p(c|s_j)
\end{split}
\end{equation}
where $p(c|s_j)$ is computed using Equation ~\ref{eq:bayes-rule}. We use the information gain to measure the degree of relevance between symptoms and conditions. For example, when the correct label was "melanoma", the first question being asked by the QA model is "what is the color of your conditioned skin ?". This is consistent with the medical knowledge confirming that the most important warning sign of melanoma is a new spot on the skin that is changing in color, size, or shape.

\section{Evaluation and Results}
We have evaluated our pipeline using 300 annotated test images which represents the only dataset with available descriptions that we used for the QA step. The evaluation metric chosen to compare the system before and after the QA model is the top-K accuracy, which is well known in medical imaging scenarios to successfully assess differential diagnosis cases. Table ~\ref{results} shows how the combination of the CNN and the QA model ameliorate the rank and the probability of the correct condition compared to the performance of each one of them separately. 
\setlength{\tabcolsep}{0.8em} 
\begin{table}[h]
\caption{The CNN and QA individual and combined performance}
\label{results}
\centering
\renewcommand{\arraystretch}{1.2}
\begin{tabular}{c|c|c|c|}
\cline{2-4}
                            & \textit{CNN model} & \textit{QA model} & \textit{CNN + QA model} \\ \hline
\multicolumn{1}{|c|}{Top-1} & 49.75 $\mypm$ 0.2\%   &  29.41\% & \textbf{57.64$\mypm$ 0.3\%}\\ \hline
\multicolumn{1}{|c|}{Top-2} & 63.55 $\mypm$ 0.6\%   &  33.21\% & \textbf{73.01$\mypm$ 0.5\%}\\ \hline
\multicolumn{1}{|c|}{Top-3} & 80.29 $\mypm$ 0.5\%   &  39.44\% & \textbf{85.62 $\mypm$ 0.5\%}\\ \hline
\end{tabular}
\end{table}
\\Here we show fivefold cross-validation classification accuracy for the CNN model. In each fold, a different fifth of the dataset is used for validation, with the rest of the dataset used for training. Reported values are the mean and standard deviation of the validation accuracy across all $n=5$ folds. The results of the QA model are separately obtained without using the condition distribution from the CNN, meaning the initial condition distribution was uniform.\\
For both the QA model separately and the combined CNN and QA models, we fix the maximum number of questions to 10 and we stop the QA process once one of the 9 condition probabilities exceeds the threshold of 95\%.\\
These results demonstrate that the proposed combination of the CNN and QA models can significantly improve the classification performance across a multiclass classification task. This can be done by utilizing medical knowledge of the relationships between conditions and symptoms, which ultimately makes the skin classification task more comparable to that of a real physician office visit, during which they are able to ask follow-up questions to narrow down the diagnosis.


\section{Conclusion}
The results of the combined CNN and QA models combined demonstrate that the proposed pipeline can effectively exploit the additional description in calibrating the final predictions across the nine skin conditions considered. It would be possible to extend this methodology to designing a study in which our CNN-QA pipeline would first be used to calibrate the CNN outputs prior to feeding the QA output to a mixture of expert networks, where each one is trained on specific conditions. We are currently investigating such an option.




\subsubsection*{Acknowledgements}
We thank Adrià Romero López, Albert Jimenez Sanfiz and the Triage team for assisting with infrastructure and evaluation, as well as with providing feedback and helpful discussions.

%

\bibliographystyle{unsrt}
\bibliography{refs}
\end{document}